\documentclass[lettersize,journal]{IEEEtran}

\ifCLASSINFOpdf
\else
   \usepackage[dvips]{graphicx}
\fi
\usepackage{url}

\usepackage{graphicx}

\begin{document}

\title{Theoretical Insight into Batch Normalization: Data Dependant Auto-Tuning of Regularization Rate}

\author{Lakshmi Annamalai, and Chetan Singh Thakur, \IEEEmembership{Senior Member, IEEE}
}

\maketitle

\begin{abstract}

Batch normalization is widely used in deep learning to normalize intermediate activations. Deep networks suffer from notoriously increased training complexity, mandating careful initialization of weights, requiring lower learning rates, etc. These issues have been addressed by Batch Normalization (\textbf{BN}), by normalizing the inputs of activations to zero mean and unit standard deviation. Making this batch normalization part of the training process dramatically accelerates the training process of very deep networks. A new field of research has been going on to examine the exact theoretical explanation behind the success of \textbf{BN}. Most of these theoretical insights attempt to explain the benefits of \textbf{BN} by placing them on its influence on optimization, weight scale invariance, and regularization. Despite \textbf{BN} undeniable success in accelerating generalization, the gap of analytically relating the effect of \textbf{BN} to the regularization parameter is still missing. This paper aims to bring out the data-dependent auto-tuning of the regularization parameter by \textbf{BN} with analytical proofs. We have posed \textbf{BN} as a constrained optimization imposed on non-\textbf{BN} weights through which we demonstrate its data statistics dependant auto-tuning of regularization parameter. We have also given analytical proof for its behavior under a noisy input scenario, which reveales the signal vs. noise tuning of the regularization parameter. We have also substantiated our claim with empirical results from the MNIST dataset experiments.

\end{abstract}

\IEEEpeerreviewmaketitle

\section{Introduction}

Deep learning architectural innovations such as convolution \cite{YannLeCunBernhard}, batch normalization, residual connections \cite{KaimingHe}, careful initialization \cite{Glorot} \cite{HeKaimingZhang} has lifted the prohibition that existed in exploring very deep networks. The most prominent innovation that has led to a phenomenal breakthrough in training deep networks has been Batch Normalization (\textbf{BN}) \cite{IoffeSzegedy}. Batch normalization has become essential component in pushing the frontiers in various Convolutional Neural Networks \cite{BarretZoph} \cite{CSzegedy} and also marked its deep root in various applications of deep learning \cite{KaimingHe} \cite{GaoHuang} \cite{DavidSilver}. 

\textbf{BN} is an additional layer that augments deep networks, which typically normalizes the activation inputs with the batch statistics $\mu$ and $\sigma$. Two trainable variables, $\gamma$, and $\beta$, are also introduced that capture the scaling and shifting operations required to be performed on the normalized activation inputs.

A number of orthogonal normalizers such as Layer normalization \cite{JimmyLei}, instance normalization \cite{DmitryUlyanov}, divisive normalization \cite{MengyeRen}, Group normalization \cite{YuxinWu}, weight normalization \cite{TimSalimans}, weight normalization with translated ReLU \cite{SitaoXiang}, decorrelated batch normalization \cite{LeiHuanDaweiYang}, iterative normalization \cite{MinhyungChoaaehyungLee}, differentiable dynamic normalization \cite{PingLuoJiaminRen} has been introduced. An alternative line of research \cite{HaninBandRolnickD} \cite{BrockADeSandSmith} \cite{ZhangHDauphinYNandMa} \cite{ShaoJHuKWangCXue} \cite{DeSandSmithS}  seeks to train deep networks without \textbf{BN}. Despite all these efforts, batch normalization still stays as milestone technique in training deep networks with its indisputable performance boost \cite{HaninBandRolnickD} \cite{MarkSandlerAndrewHoward} \cite{BarretZophVijayVasudevan} \cite{JieHuLiShen} \cite{AndrewGHowardMenglong}. 

Two related veins of research that have emerged for \textbf{BN} are i) Branch which studies the theory behind the benefits of \textbf{BN} and ii) Branch, which seeks to find ways to improvise and solve the current issues of \textbf{BN}. Research contribution towards improvising performance of \textbf{BN} mostly concentrate on alleviating its dependency on batch size \cite{KHeXZhang} \cite{SIoffe} \cite{CPengTXiao} \cite{YongGuoQingyaoWu}. \cite{NilsBjorck} also came up with a new flavor of \textbf{BN}, which could be applied to time sequence models such as Recurrent Neural Network (RNN).

Despite the undeniable vein of research in analyzing the benefits of \textbf{BN}, most of it has been aimed at theoretically explaining its success in terms of improved gradient updates, smoothness in the optimization landscape, larger learning rates, and scale invariance of weights. Our theoretical analysis, on the other hand, identifies a new perspective for the success of \textbf{BN} by projecting its effect on auto-tuning of the regularization parameter. 

Regularization rate choice is crucial to hitting the right balance between model simplicity and training data over-fitting. The ideal value of the regularization rate is data dependent and choosing a safe regularization parameter is still a hot topic of ongoing research in machine learning. In deep learning, cross-validation is the most well-known technique used to select optimal hyperparameter, including regularization rate, whereas \textbf{BN} solves this issue by automatically estimating the optimal regularization rate to be applied on the non-\textbf{BN} weights for the given data.

In this work, we have brought out two different perspectives on the \textbf{BN} based auto-tuning of the regularization parameter. i) By defining \textbf{BN} as a constrained optimization imposed on non-\textbf{BN} weights, we have theoretically brought out that \textbf{BN} imposes optimal data dependant regularization effect on the estimated non-\textbf{BN} weights while making sure that the estimated \textbf{BN} weights do not deviate from the former. The optimality in the regularization is brought out by enabling the node to estimate the regularization rate from the training data statistics. ii) By analyzing noise-injected input data, we have analytically shown that \textbf{BN} varies the regularization parameter to signal vs. noise variance.

\section{Related Work}

Several recent works \cite{GregYang} \cite{ZhiyuanLi} \cite{JonasKohler} have tried to approach BatchNorm with the angle of theoretical explanation, especially theoretical analysis of how BN favors i) optimization by minimizing the Hessian norm, by making weights scale invariant and ii) generalization by allowing higher learning rates and by implicitly regularizing the network.  

\cite{IoffeSzegedy} proposed a mechanism known as Batch Normalization (\textbf{BN}) to eliminate the critical problem of change in the distribution of non-linearity inputs, thus making the training process well behaved. It has been argued in \cite{IoffeSzegedy} that BN reduces the internal covariance shift. It was proved in \cite{Lipton} \cite{Santurkar} with a strong hypothesis that convergence of BN can not be explained by internal covariance shift but by smoothening of the landscape. \cite{Santurkar} demonstrated the effect of \textbf{BN} on the stability of the loss function in terms of its Lipschitness. They have shown that the gradient magnitude of the loss function with \textbf{BN} $\Vert{\delta{L_{BN}}}\Vert$ is relatively flat as compared to that of the one without \textbf{BN}. Towards second-order optimization properties, they have proved that loss Hessian (second order term of Taylor expansion) is reduced, thus making gradient (first order term) more predictive. \cite{JonasKohler} proved that \textbf{BN} accelerates optimization by decoupling the length and directional components of the weight vector, which is nothing but non-linear re-parametrization of the weight space.

\cite{AriMorcos} empirically showed that the networks trained with \textbf{BN} encourage good generalization by making the network less sensitive to individual activations/feature maps, which is commonly observed in 'memorizing networks.' \cite{NilsBjorck} argues that BN improves generalization by allowing a higher learning rate. It was suggested that noise in SGD is proportional to $\frac{\lambda^2}{|B|}$, where $\lambda$ is the learning rate and $|B|$ is the batch size. \textbf{BN} allows a larger learning rate, which in turn increases the noise in optimization and hence biases the network towards good generalization. It also proves that BN has the luxury of higher learning rates. Without BN, the activation overgrows at deep layers, with the network output exploding and loss diverging. BN mitigates this phenomenon by standardizing the activation at each layer, thus making sure that the network always lands in a safe plateau of optimization function irrespective of the higher learning rate. \cite{PingLuo} says that \textbf{BN} provides explicit regularization by encouraging the participation of all neurons, reducing the kurtosis of the input distribution and the correlation between neurons.

\cite{MinhyungCho} showed that \textbf{BN} makes the network invariant to linear scaling of weights, i.e. \textbf{BN}$(w^T*x)=$\textbf{BN}$(u^T*x)$, where $u=\frac{w}{\Vert{w}\Vert}$ \cite{GuodongZhang}. \cite{EladHoffer} finds the mechanism through which \textbf{BN} makes the optimization invariant to weight scale, by proving that the increment in weight $w_{t+1}-w_{t}$ is inversely proportional to the norm of the weight $\Vert{w_t}\Vert^2_2$ for linear layer followed by \textbf{BN} and smaller learning rate.

\cite{YongqiangCai} provided analysis of \textbf{BN} on an ordinary least square problem that converges for any learning rate irrespective of the spectral radius of the Hessian matrix. In contrast, in gradient descent without \textbf{BN}, the range of learning rate is $0 \geq \lambda \leq {\frac{2}{\lambda_{max}}}$, where $\lambda_{max}$ is the maximum learning rate. \cite{Arora} showed that the optimization minimization of the networks with \textbf{BN} does not require learning rate tuning. Full batch gradient descent converges to a first order stationary point at the rate of $T^{(-\frac{1}{2})}$, no matter what the learning rate has been set.

\cite{DavidBalduzzi} states the effect of \textbf{BN} on neuronal activity and concludes that it enables efficient usage of rectifier non-linearities.

\section{Theoretical Interpretation of Batch Normalization}

This section analyses the working of \textbf{BN} in a single perceptron, which forms the building block of deep networks. The computation of \textbf{BN} can be written as follows,

\begin{equation}
y=\gamma\frac{\mathbf{x}^t\mathbf{w}-\mu}{\sigma}+\beta
\label{eq:1}
\end{equation}

Where, $\mathbf{x},\mathbf{w}\in{\emph{R}^d}$ are the input and weight vector respectively, $y$ is the batch normalized output. $\mu$ and $\sigma$ are the mean and standard deviation of $\mathbf{x}$ projected onto $\mathbf{w}$, respectively. $\gamma$ and $\beta$ are trainable scale and shift parameters of \textbf{BN}.

The zero mean weak assumption is made on input data $\mathbf{x}$ such that $\mathbf{\emph{E}}(\mathbf{x})=0$. As given in \cite{JonasKohler}, Eq. \ref{eq:1} can be written as follows (omitting $\beta$)

\begin{equation}
y=\gamma\frac{\mathbf{x}^t\mathbf{w}}{(\mathbf{w}^t\mathbf{R}\mathbf{w})^{\frac{1}{2}}}
\label{eq:2}
\end{equation}

Where, $\mathbf{R}=\mathbf{\emph{E}}[\mathbf{x}\mathbf{x}^t]$ is the symmentric positive definite covariance matrix of $\mathbf{x}$. This can be written as $y=\mathbf{x}^t\mathbf{w}_{bn}$, where, $\mathbf{w}_{bn}=g\frac{\mathbf{w}}{(\mathbf{w}^t\mathbf{R}\mathbf{w})^{\frac{1}{2}}}$, is the weight of the perceptron with \textbf{BN} operation preceeding the application of activation.

\begin{figure}
\centerline{\includegraphics[height=3in,width=2in,keepaspectratio=true]{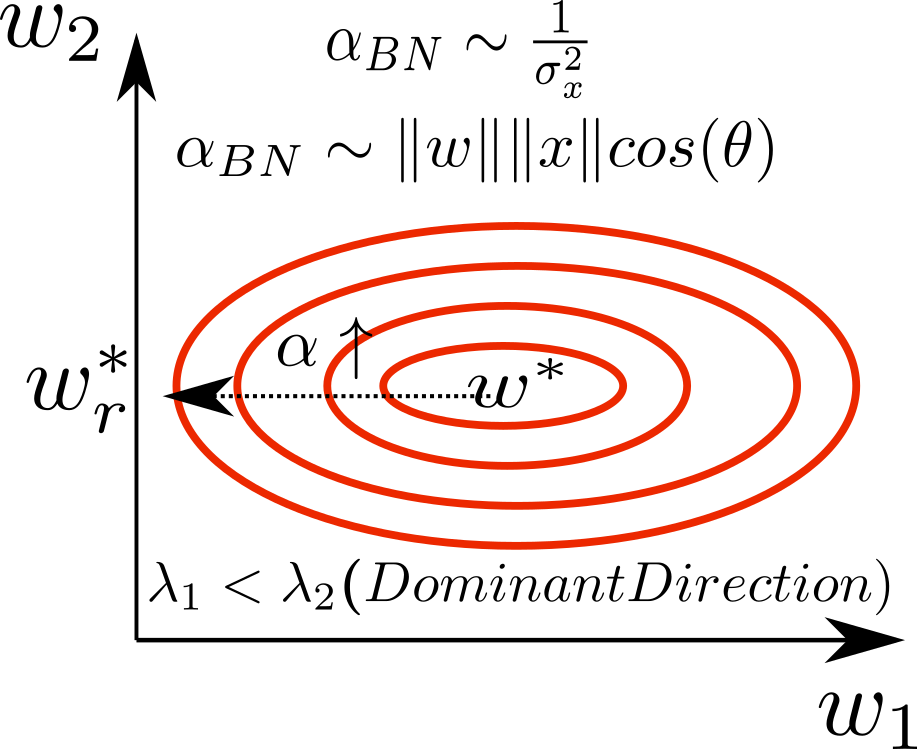}}
\caption{Two dimensional loss landscape of $w$ with optimal value at $w^*$. The parameter $\alpha$ determines the intensity of regularization induced, retaining only significant directions corresponding to larger eigenvalues ($\lambda_2$ in the example given). With increase in $\alpha$, optimal $w$ converges towards $w_r$. However, hyperparameter tuning of $\alpha$ is critical to achieving good performance. \textbf{BN} alleviates this major hurdle by estimating optimal $\alpha_{BN}$ from the data $x$ and non-\textbf{BN} weight vector. Models and data with larger values are more prone to classification error even with small perturbations in $x$ and $w$, thus resulting in increased generalization error. In addition, regularization should also depend on the $\theta$ between $x$ and $w$. When the data points are close to the decision boundary, the intensity of regularization should be minimal as there is a high probability that the regularized weights may move the data points to the wrong side of the classification boundary. By introducing proportional relation between the regularization parameter ${\Vert{w}\Vert, \Vert{x}\Vert}$ and $\cos{\theta}$, $\alpha_{BN}$ improves the generalization capability. The network's tendency to memorize the training data decreases with increased data variance $\sigma_x^2$. $\alpha_{BN} \sim \sigma_x^2$ signifies the tuning of regularization with respect to data variance.}
\label{fig:1}
\end{figure}

\subsection{\textbf{BN} as Constrained Optimization applied on non-\textbf{BN} Weights}

\textbf{BN} defines the following optimality criterion: Minimize the weighted norm of $\mathbf{w}_{bn}$ while keeping the projection of the \textbf{BN} weight vector onto the non-\textbf{BN} weight vector fixed. 

\begin{equation}
\arg\min_{\mathbf{w}_{bn}}[{\mathbf{\emph{E}}\Vert{\mathbf{w}^t\mathbf{x}}\Vert}^2_2]^\frac{1}{2}\mathbf{w}_{bn}^t\mathbf{w}_{bn}
\end{equation} 

with the constraint,

\begin{equation}
\mathbf{w}_{bn}^t\mathbf{w}=1
\end{equation}

Solving the constrained minimization by the method of Lagrange multipliers, we have

\begin{equation}
arg\min_{\mathbf{w}_{bn}}F(w;\alpha_b)=[{\mathbf{\emph{E}}\Vert{\mathbf{w}^t\mathbf{x}}\Vert}_2^2]^\frac{1}{2}\mathbf{w}_{bn}^t\mathbf{w}_{bn}-\alpha_b[\mathbf{w}_{bn}^t\mathbf{w}-1]
\label{eq:3}
\end{equation}

On simplification of the first term,

\begin{eqnarray}
&[{\mathbf{\emph{E}}{(\mathbf{w}^t\mathbf{x})}^t(\mathbf{w}^t\mathbf{x})}]^\frac{1}{2}\mathbf{w}_{bn}^t\mathbf{w}_{bn} \\
=&\left[\mathbf{w}^t\mathbf{\emph{E}}(\mathbf{x}\mathbf{x}^t)\mathbf{w}\right]^\frac{1}{2}\mathbf{w}_{bn}^t\mathbf{w}_{bn} \\
=&\left[\mathbf{w}^t\mathbf{R}\mathbf{w}\right]^\frac{1}{2}\mathbf{w}_{bn}^t\mathbf{w}_{bn}
\end{eqnarray}

Substituting this simplified form in Eq. \ref{eq:3}, $arg\min_{\mathbf{w}_{bn}}F(w;\alpha_b)$ becomes

\begin{equation}
\left[\mathbf{w}^t\mathbf{R}\mathbf{w}\right]^\frac{1}{2}\mathbf{w}_{bn}^t\mathbf{w}_{bn}-\alpha_b[\mathbf{w}_{bn}^t\mathbf{w}-1]
\end{equation}

Differentiating with respect to $\mathbf{w}_{bn}$ and equating it to zero, we get

\begin{equation}
\left[\mathbf{w}^t\mathbf{R}\mathbf{w}\right]^\frac{1}{2}\mathbf{w}_{bn}-\alpha_b[\mathbf{w}] = 0
\end{equation}

Therefore,

\begin{equation}
\mathbf{w}_{bn}=\alpha_b\frac{\mathbf{w}}{(\mathbf{w}^t\mathbf{R}\mathbf{w})^{\frac{1}{2}}}
\label{eq:4}
\end{equation}

Eq. \ref{eq:4} turns out to be similar \textbf{BN} normalized output as given in Eq. \ref{eq:2} when $\alpha_b=\gamma$.

\subsubsection{Benefits of Minimizer and Constraint}

The constraint ensures that the information learned by the non-\textbf{BN} weight vector is preserved and no distortion happens by the introduction of \textbf{BN}.

The minimizer is a form of regularization function defined over $\mathbf{w}_{bn}$ (Fig. \ref{fig:1}). The relation between unregularized and $\mathbf{\emph{L}}_2$ regularized weights $\mathbf{w}_r$ is given as $\mathbf{w}_r=\mathbf{Q}(\mathbf{\Lambda}+\alpha\mathbf{I})\mathbf{\Lambda}\mathbf{Q}^t\mathbf{w}$, where $\mathbf{Q},\mathbf{\Lambda}$ are eigenvector and eigenvalue matrices of the Hessian matrix of weight vector respectively and $\alpha$ is the Langrangian parameter which determines the scaling applied to the regularization term. Effectively, $i^{th}$ component of weight vector is rescaled by $\frac{\lambda_i}{\lambda_i+\alpha}$, hence retaining only dominant directions. However, the number of dominant directions retained is influenced by $\alpha$, which is generally a hyperparameter in optimization, whereas in the \textbf{BN} minimization function $\alpha$ is obtained from $[{\mathbf{\emph{E}}\Vert{\mathbf{w}^t\mathbf{x}}\Vert}^2_2]^\frac{1}{2}$. 

When $\mathbf{w}^t\mathbf{x}$ is higher, even a small perturbation in input $\mathbf{x+\varepsilon} \approx \mathbf{x}$ will cause huge deviation in the output $\mathbf{w}^t(\mathbf{x+\varepsilon}) \gg \mathbf{w}^t\mathbf{x}$, thus increasing the variance in the model. This effect is mitigated by having the term $[{\mathbf{\emph{E}}\Vert{\mathbf{w}^t\mathbf{x}}\Vert}_2^2]^\frac{1}{2}$ as $\alpha$ in the $\mathbf{w}_{bn}$ regularization function. The higher the value of $\mathbf{w}^t\mathbf{x}$, the higher the effect of $\alpha$, thus resulting in increased regularization. 








\subsection{Signal vs. Noise Variance based Tuning of Regularization Parameter by \textbf{BN}}

From the previous section, it could be inferred that,

\begin{equation}
\mathbf{w}_{bn}=\gamma\frac{\mathbf{w}}{(\mathbf{w}^t\mathbf{R}\mathbf{w})^{\frac{1}{2}}}
\label{eq:5}
\end{equation}

$\mathbf{R}$ is a symmetric and positive definite matrix. Hence there exists a matrix $\mathbf{B}$ such that $B=\mathbf{Q}\mathbf{\Lambda}\mathbf{Q}^t$ and $\mathbf{R}=\mathbf{B}^2$. Hence Eq. \ref{eq:5} becomes

\begin{eqnarray}
\mathbf{w}_{bn}&=&\gamma\frac{\mathbf{w}}{(\mathbf{w}^t\mathbf{B}^t\mathbf{B}\mathbf{w})^{\frac{1}{2}}} \\
&=&\gamma\frac{\mathbf{w}}{{\left[{(\mathbf{B}\mathbf{w})}^t{(\mathbf{B}\mathbf{w})}\right]}^{\frac{1}{2}}}
\end{eqnarray}

$\mathbf{w}$ can be written as $w=\sum_{i=1}^{i=d}{\mathbf{q}_iw_i}$, where $\mathbf{q}_i$ is the $i^{th}$ eigen vector of the matrix $\mathbf{R}$ and $w_i$ is the $i^{th}$ element of the $d$ dimensional $\mathbf{w}$ vector. Hence, $\mathbf{B}\mathbf{w}$ becomes $\sum_{i=1}^{i=d}{\mathbf{B}\mathbf{q}_iw_i}$ which turns the equation of $\mathbf{w}_{bn}$ to as follows,

\begin{equation}
\mathbf{w}_{bn}=\gamma\frac{\mathbf{w}}{{\left[\sum_{i=1}^{i=d}{\lambda_i}w_i^2\right]}^\frac{1}{2}}
\label{eq:6}
\end{equation}

Let's see how the modified \textbf{BN} weight vector regulates overfitting based on the noise in the input data. The input data corrupted with gaussian noise $\xi\sim{\textbf{N}(0,\sigma_n^2)}$ such that $\mathbf{x}_n=\mathbf{x}+\mathbf{\xi}$. The relation between the mean square error between the true output $\mathbf{y}$ and the predicted outputs (with noise: $\mathbf{y}^p_n$, without noise: $\mathbf{y}^p$) is given as follows

\begin{equation}
\mathbf{E}{\left[{\mathbf{y}^p_n-\mathbf{y}}\right]}^2=\mathbf{E}{\left[{\mathbf{y}^p-\mathbf{y}}\right]}^2+\sigma^2\sum_{i=1}^{i=d}w_i^2
\end{equation}

The second term in the above equation is the regularization term introduced by the noise induced in the input. The regularization parameter is proportional to $\sigma^2$. Substituting $\mathbf{w}_{bn}$ in the place of $\mathbf{w}$ as per Eq. \ref{eq:6}, we get,

\begin{equation}
\mathbf{E}{\left[{\mathbf{y}^p_n-\mathbf{y}}\right]}^2=\mathbf{E}{\left[{\mathbf{y}^p-\mathbf{y}}\right]}^2+\sigma^2\sum_{i=1}^{i=d}{\gamma^2\frac{\mathbf{w}^2}{\sum_{j=1}^{j=d}{\lambda^n_j}w_j^2}}
\end{equation}

Where, $\lambda^n_j=\lambda_j+\sigma^2$ is the $j^{th}$ eigenvalue of noisy covariance matrix. (The data covariance matrix of noisy data is given as $\mathbf{R}_n=\mathbf{R}+\sigma^2\mathbf{I}$ as the data and noise are uncorrelated). 

\begin{equation}
\mathbf{E}{\left[{\mathbf{y}^p_n-\mathbf{y}}\right]}^2=\mathbf{E}{\left[{\mathbf{y}^p-\mathbf{y}}\right]}^2+\sigma^2\gamma^2\sum_{i=1}^{i=d}{\frac{\mathbf{w}^2}{{\left[\sum_{j=1}^{j=d}{(\lambda_j+\sigma^2})w_j^2\right]}}}
\end{equation}

Assuming all $\lambda_j$ to be equal, the above equation becomes

\begin{equation}
\mathbf{E}{\left[{\mathbf{y}^p_n-\mathbf{y}}\right]}^2=\mathbf{E}{\left[{\mathbf{y}^p-\mathbf{y}}\right]}^2+\frac{\sigma^2\gamma^2}{(\lambda+\sigma^2)}\sum_{i=1}^{i=d}{\frac{\mathbf{w}^2}{{\left[\sum_{j=1}^{j=d}w_j^2\right]}}}
\end{equation}

When $\sigma >> \lambda$, $\frac{\sigma^2\gamma^2}{(\lambda+\sigma^2)}=\gamma^2$ and when $\sigma << \lambda$, $\frac{\sigma^2\gamma^2}{(\lambda+\sigma^2)}=\frac{\gamma^2}{\lambda}$, thus making the regularization parameter inversely proportional to signal variance. Whenever the variance in the signal is high, it is challenging for the model to memorize the data, reducing the need for regularization. \ textbf{BN} weights achieve this by auto-tuning the regularization parameter based on the signal and noise variance.


\section{Experiments and Results}

We investigated the proposed theoretical justification of \textbf{BN} on a Multi-Layer Perceptron (MLP), evaluated on the MNIST dataset. The aim was rather to test the proposed hypothesis regarding the data dependant regularization capability of \textbf{BN}. We designed two sets of experiments, where the auto-tuning effect of the regularization parameter of \textbf{BN} and signal-to-noise ratio dependant regularization of \textbf{BN} has been tested. 

\subsection{$[{\mathbf{\emph{E}}\Vert{\mathbf{w}^t\mathbf{x}}\Vert}^2_2]^\frac{1}{2}$: Regularization Parameter of BN applied Weights}

To verify the effect of $[{\mathbf{\emph{E}}\Vert{\mathbf{w}^t\mathbf{x}}\Vert}^2_2]^\frac{1}{2}$ introduced by \textbf{BN} on regularization of weights, we have considered the classification problem of predicting the digit class from MNIST dataset. The motive of the experiment is to analyse the dependence between regularization of \textbf{BN} and $[{\mathbf{\emph{E}}\Vert{\mathbf{w}^t\mathbf{x}}\Vert}^2_2]^\frac{1}{2}$, not to achieve state-of-the-art performance. Hence, we have used a shallow network of one hidden layer of $300$ nodes (with ReLu activation) followed by a dense classifier layer (with SoftMax activation) of $10$ nodes. \textbf{BN} is applied right after every dense layer. We have trained it with a batch size of $128$, categorical cross-entropy loss, and Adam optimizer with a learning rate of $0.1$. The weights are initialized with small values drawn from the Gaussian Distribution. 

To study the regularization introduced by \textbf{BN}, we demonstrate the impact of \textbf{BN} on the norm of the weight. The higher the norm, the lesser the regularization. To analyze the effect of $\mathbf{w}^t\mathbf{x}$ as a regularization parameter, we have applied scaling to $\mathbf{x}$ rather than to $\mathbf{w}$. 

\begin{table}[!ht]
\centering
\begin{tabular}{c c c c} 
\hline
\textbf{Data Amplitude} & \textbf{$L_1$ Norm} & \textbf{$L_2$ Norm} & $\alpha ({\times}10^8)$ \\ \hline
2 & 0.7210 & 0.0322 &  0.92\\
3 & 0.3003 & 0.0134 & 2.2\\
4 & 0.1717 & 0.0077 & 3.9\\
5 & 0.1096 & 0.0049 & 5.9\\
6 & 0.0793 & 0.0035 & 8.3\\ \hline

\end{tabular}
\caption{$\alpha$ (regularization rate) and Mean of $L_1$ and $L_2$ norm of weight vector across $300$ hidden nodes with respect to scaling applied to the data. The norm of the weight vector is observed to increase with the scaling factor of data, thus illustrating the increase in the regularization intensity. It shows that the higher the scaling of data, the higher the $\alpha$}
\label{table:1}
\end{table}

Table. \ref{table:1} shows the mean of $L_1$ and $L_2$ norms of the weight across $300$ hidden nodes for various scaling factors (varied from $2$ to $6$). It could be observed that the norm of the weights decreases with an increase in the scaling factor, thus offering regularization auto-tuning capability to \textbf{BN} based on $\mathbf{w}^t\mathbf{x}$.

\begin{figure}
\centerline{\includegraphics[width=\columnwidth]{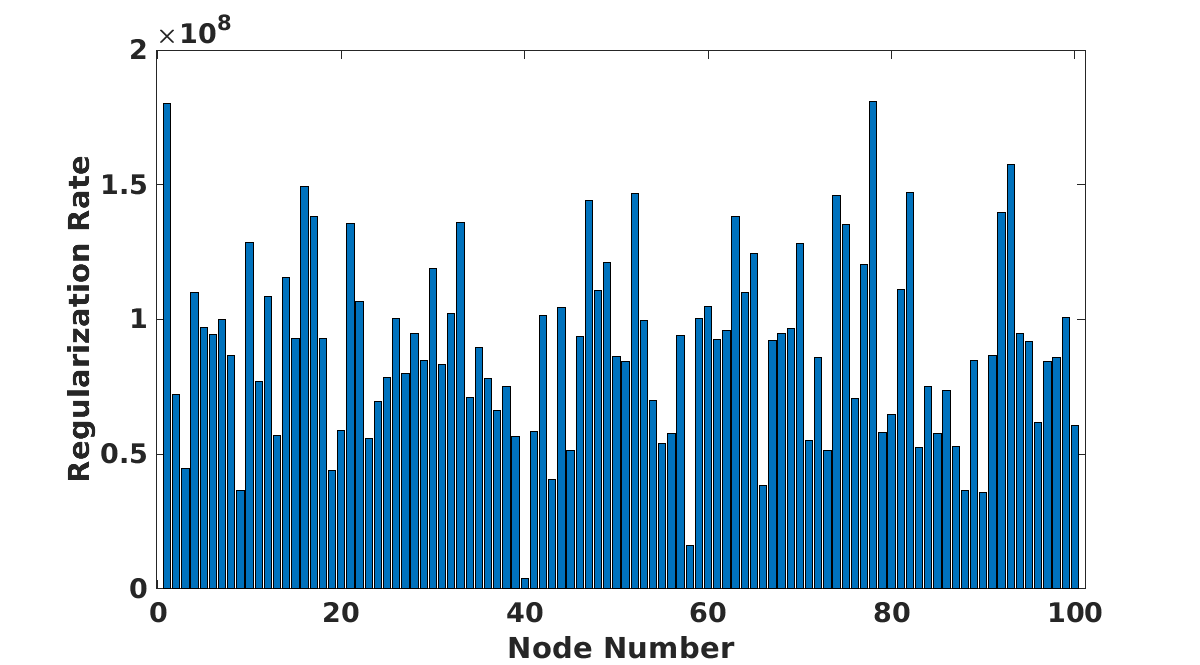}}
\caption{$\alpha$ (regularization rate) estimated by \textbf{BN} operation across $300$ hidden nodes. It shows that $\alpha$ is variable across nodes.}
\label{fig:2}
\end{figure}

Table. \ref{table:1} reports the mean of the $\alpha$ introduced by \textbf{BN} across $300$ hidden nodes. We could infer that \textbf{BN} not only makes $\alpha$ dependant on the scaling applied to the data but also optimizes it to a better value suitable for each node (Fig. \ref{fig:2}), unlike non-\textbf{BN} networks, where single $\alpha$ is applied across all nodes. We found that the optimal estimation of $\alpha$ for every node is surprising.

\subsection{Signal vs. Noise Variance based Regularization Effect of \textbf{BN}}

Next, we propose the following experimental setup to disentangle the adaptive regularization tuning effect of \textbf{BN} based on $\lambda$ vs. $\sigma$. To study the impact of noise in the input, we propose the following experiment: We injected additive noise sampled from i.i.d Gaussian distribution to the MNIST dataset. Note that this noise injection perturbs the input Signal to Noise ratio. We then measure the influence of this deliberately introduced SNR variation on regularization introduced by \textbf{BN}. 

Fig. \ref{table:2} visualizes the mean of the $L_1$ and $L_2$ norm of weights across $300$ hidden nodes under different SNR conditions. It can be observed that the regularization difference between high and low SNR is existent, hence making it apparent that the intensity of regularization is directly connected to the ratio $\frac{\sigma^2\gamma^2}{(\lambda+\sigma^2)}$.

\begin{table}[!ht]
\centering
\begin{tabular}{c c c} 
\hline
\textbf{SNR} & \textbf{$L_1$ Norm} & \textbf{$L_2$ Norm} \\ \hline
100 & 2.8823 & 0.1286 \\
10 & 2.5811 & 0.1152 \\
1 & 0.6090 & 0.0272 \\
0.5 & 0.3646 & 0.0163 \\ \hline

\end{tabular}
\caption{Mean of $L_1$ and $L_2$ norm of the weights of $300$ hidden nodes under different SNR. Weight norm decreases with increased SNR, thus proving that \textbf{BN} makes the regularization parameter inversely proportional to SNR.}
\label{table:2}
\end{table}

\section{Conclusion}

We have given a different theoretical approach to analyzing the advantages provided by batch normalization. It is based on the premise that the regularization parameter, known to be a hyper-parameter in non-\textbf{BN} networks, has been estimated automatically by the \textbf{BN}. We have shown that \textbf{BN} acts as a constrained optimization applied on non-\textbf{BN} weights, thus affording auto-tuning of regularization parameters. Note that we have also given proof that the regularization parameter tuning of \textbf{BN} is SNR dependent. To evaluate the proposed hypothesis, we have conducted experiments on the MNIST dataset. The results of the conducted experiments substantiate the proposed theoretical analysis. In future work, we are interested in analyzing this regularization auto-tuning effect of \textbf{BN} in deep networks, which is still an open research problem.


\begin{thebibliography}{34}

\bibitem{Glorot}
Glorot, Xavier and Bengio, Yoshua. Understanding the difficulty of training deep feedforward neural networks. In AISTATS, 2010

\bibitem{HeKaimingZhang}
He, Kaiming, Zhang, Xiangyu, Ren, Shaoqing, and Sun, Jian. Delving Deep into Rectifiers: Surpassing Human-Level Performance on ImageNet Classification. In ICCV, 2015

\bibitem{YannLeCunBernhard}
Yann LeCun, Bernhard E Boser, John S Denker, Donnie Henderson, Richard E Howard, Wayne E Hubbard, and Lawrence D Jackel. Handwritten digit recognition with a back-propagation network. In Advances in neural information processing systems, pp. 396–404, 1990.

\bibitem{KaimingHe}
Kaiming He, Xiangyu Zhang, Shaoqing Ren, and Jian Sun. Deep Residual Learning for Image Recognition. arXiv:1512.03385 [cs], December 2015.

\bibitem{JonasKohler}Jonas Kohler, Hadi Daneshmand, Aurelien Lucchi, Ming Zhou, Klaus Neymeyr, and Thomas Hofmann. Towards a theoretical understanding of batch normalization. arXiv preprint arXiv:1805.10694, 2018.

\bibitem{IoffeSzegedy} Ioffe, S. and Szegedy, C. Batch normalization: Accelerating deep network training by reducing internal covariate shift. In Proceedings of the 32nd International Conference on International Conference on Machine Learning - Volume 37, ICML, pages 448–456, 2015.

\bibitem{Lipton} Lipton, Z. C. and Steinhardt, J. (2018). Troubling trends in machine learning scholarship. arXiv preprint arXiv:1807.03341

\bibitem{Santurkar} Santurkar, S., Tsipras, D., Ilyas, A., and Madry, A. (2018). How does batch normalization help optimization?(no, it is not about internal covariate shift). arXiv preprint arXiv:1805.11604

\bibitem{NilsBjorck} Nils Bjorck, Carla P Gomes, Bart Selman, and Kilian Q Weinberger. Understanding batch normalization. In S. Bengio, H. Wallach, H. Larochelle, K. Grauman, N. Cesa-Bianchi, and R. Garnett (eds.), Advances in Neural Information Processing Systems 31, pp. 7705–7716. Curran Asso- ciates, Inc., 2018

\bibitem{Arora} Arora, S., Li, Z., and Lyu, K. Theoretical analysis of auto rate-tuning by batch normalization. In In- ternational Conference on Learning Representations, 2019.

\bibitem{JonasKohler} Jonas Kohler, Hadi Daneshmand, Aurelien Lucchi, Thomas Hofmann, Ming Zhou, Klaus Neymeyr. Exponential convergence rates for Batch Normalization: The power of length-direction decoupling in non-convex optimization. AISTATS, 2019

\bibitem{ZhiyuanLi} Zhiyuan Li and Sanjeev Arora, An Exponential Learning Rate Schedule For Batch Normalized Networks. ICLR, 2020

\bibitem{YongqiangCai} Yongqiang Cai, Qianxiao Li and Zuowei Shen, A Quantitative Analysis of the Effect of Batch Normalization on Gradient Descent


\bibitem{GuodongZhang} Guodong Zhang, Chaoqi Wang, Bowen Xu, and Roger Grosse. Three mechanisms of weight decay regularization. In International Conference on Learning Representations, 2019

\bibitem{MinhyungCho} Minhyung Cho and Jaehyung Lee. Riemannian approach to batch normalization. In I. Guyon, U. V. Luxburg, S. Bengio, H. Wallach, R. Fergus, S. Vishwanathan, and R. Garnett (eds.), Advances in Neural Information Processing Systems 30, pp. 5225–5235. Curran Associates, Inc., 2017

\bibitem{EladHoffer} Elad Hoffer, Ron Banner, Itay Golan, and Daniel Soudry. Norm matters: efficient and accurate normalization schemes in deep networks. In S. Bengio, H. Wallach, H. Larochelle, K. Grauman, N. Cesa-Bianchi, and R. Garnett (eds.), Advances in Neural Information Processing Systems 31, pp. 2164–2174. Curran Associates, Inc., 2018

\bibitem{GregYang} Greg Yang, Jeffrey Pennington, Vinay Rao, Jascha Sohl-Dickstein, and Samuel S. Schoenholz. A mean field theory of batch normalization. In International Conference on Learning Representations, 2019

\bibitem{PingLuo} Ping Luo, Xinjiang Wang, Wenqi Shao, and Zhanglin Peng. Towards understanding regularization in batch normalization. In International Conference on Learning Representations, 2019

\bibitem{AriMorcos} Ari Morcos, David GT Barrett, Neil C Rabinowitz, and Matthew Botvinick. On the importance of single directions for generalization. In Proceeding of the International Conference on Learning Representations, 2018

\bibitem{DavidBalduzzi} David Balduzzi, Marcus Frean, Lennox Leary, JP Lewis, Kurt Wan-Duo Ma, and Brian McWilliams. The shattered gradients problem: If resnets are the answer, then what is the question? In Proceedings of the 34th International Conference on Machine Learning-Volume 70, pp. 342–350. JMLR. org, 2017

\bibitem{JimmyLei}
Jimmy Lei Ba, Jamie Ryan Kiros, and Geoffrey E Hinton. Layer normalization. arXiv:1607.06450, 2016

\bibitem{DmitryUlyanov}
Dmitry Ulyanov, Andrea Vedaldi, and Victor Lempitsky. Instance normalization: The missing ingredient for fast stylization. arXiv preprint arXiv:1607.08022, 2016

\bibitem{MengyeRen}
Mengye Ren, Renjie Liao, Raquel Urtasun, Fabian H Sinz, and Richard S Zemel. Normalizing the normalizers: Comparing and extending network normalization schemes. arXiv preprint arXiv:1611.04520, 2016

\bibitem{TimSalimans}
Tim Salimans and Diederik P Kingma. Weight normalization: A simple reparameterization to accelerate training of deep neural networks. In Advances in Neural Information Processing Systems, pp. 901–909, 2016.

\bibitem{SitaoXiang}
Sitao Xiang and Hao Li. On the effects of batch and weight normalization in generative adversarial networks. stat, 1050:22, 2017

\bibitem{BarretZoph}
Barret Zoph, Vijay Vasudevan, Jonathon Shlens, and Quoc V Le. Learning transferable architectures for scalable image recognition

\bibitem{CSzegedy}
C. Szegedy, V. Vanhoucke, S. Ioffe, J. Shlens, and Z. Wojna. Rethinking the inception architecture for computer vision. In CVPR, 2016

\bibitem{KaimingHe}
Kaiming He, Xiangyu Zhang, Shaoqing Ren, and Jian Sun. Deep residual learning for image recognition. In Proceedings of the IEEE conference on computer vision and pattern recognition, pages 770–778, 2016.

\bibitem{GaoHuang}
Gao Huang, Zhuang Liu, Kilian Q Weinberger, and Laurens van der Maaten. Densely connected convolutional networks. In Proceedings of the IEEE conference on computer vision and pattern recognition, volume 1, page 3, 2017

\bibitem{DavidSilver}
David Silver, Julian Schrittwieser, Karen Simonyan, Ioannis Antonoglou, Aja Huang, Arthur Guez, Thomas Hubert, Lucas Baker, Matthew Lai, Adrian Bolton, et al. Mastering the game of go without human knowledge. Nature, 550(7676):354, 2017

\bibitem{YuxinWu}
Yuxin Wu, Kaiming He, Group Normalization.

\bibitem{KHeXZhang}
K. He, X. Zhang, S. Ren, and J. Sun. Deep residual learning for image recognition. In CVPR, 2016

\bibitem{SIoffe}
S. Ioffe. Batch renormalization: Towards reducing minibatch dependence in batch-normalized models. In NIPS, 2017

\bibitem{CPengTXiao}
C. Peng, T. Xiao, Z. Li, Y. Jiang, X. Zhang, K. Jia, G. Yu, and J. Sun. MegDet: A large mini-batch object detector. In CVPR, 2018

\bibitem{YongGuoQingyaoWu}
Yong Guo, Qingyao Wu, Chaorui Deng, Jian Chen, and Mingkui Tan. Double forward propagation for memorized batch normalization. In Thirty Second AAAI Conference on Artificial Intelligence, 2018.

\bibitem{LeiHuanDaweiYang}
Lei Huang, Dawei Yang, Bo Lang, and Jia Deng. Decorrelated batch normalization. In Proceedings of the IEEE Conference on Computer Vision and Pattern Recognition, pp. 791–800, 2018.

\bibitem{MinhyungChoaaehyungLee}
Minhyung Cho and Jaehyung Lee. Riemannian approach to batch normalization. In Advances in Neural Information Processing Systems, pp. 5225–5235, 2017.

\bibitem{PingLuoJiaminRen}
Ping Luo, Jiamin Ren, and Zhanglin Peng. Differentiable learning-to-normalize via switchable normalization. In International Conference on Learning Representations, 2019

\bibitem{HaninBandRolnickD}
Hanin, B. and Rolnick, D. How to start training: The effect of initialization and architecture. In Advances in Neural Information Processing Systems, pp. 571–581, 2018.

\bibitem{BrockADeSandSmith}
Brock, A., De, S., and Smith, S. L. Characterizing signal propagation to close the performance gap in unnormalized resnets. In 9th International Conference on Learning Representations, ICLR, 2021.

\bibitem{ZhangHDauphinYNandMa}
Zhang, H., Dauphin, Y. N., and Ma, T. Fixup initialization: Residual learning without normalization. arXiv preprint arXiv:1901.09321, 2019a

\bibitem{ShaoJHuKWangCXue}
Shao, J., Hu, K., Wang, C., Xue, X., and Raj, B. Is normalization indispensable for training deep neural network? Advances in Neural Information Processing Systems, 33, 2020.

\bibitem{DeSandSmithS}
De, S. and Smith, S. Batch normalization biases residual blocks towards the identity function in deep networks. Advances in Neural Information Processing Systems, 33, 2020

\bibitem{AndrewBrockSohamDe}
Andrew Brock, Soham De, Samuel L. Smith and Karen Simonyan, High-Performance Large-Scale Image Recognition Without Normalization.

\bibitem{HaninBandRolnickD}
Hanin, B. and Rolnick, D. How to start training: The effect of initialization and architecture. In Advances in Neural Information Processing Systems, pp. 571–581, 2018

\bibitem{MarkSandlerAndrewHoward}
Mark Sandler, Andrew Howard, Menglong Zhu, Andrey Zhmoginov, and Liang-Chieh Chen. Mo- bilenetv2: Inverted residuals and linear bottlenecks. In Proceedings of the IEEE Conference on Computer Vision and Pattern Recognition, pp. 4510–4520, 2018

\bibitem{BarretZophVijayVasudevan}
Barret Zoph, Vijay Vasudevan, Jonathon Shlens, and Quoc V Le. Learning transferable architectures for scalable image recognition. In Proceedings of the IEEE conference on computer vision and pattern recognition, pp. 8697–8710, 2018.

\bibitem{JieHuLiShen}
Jie Hu, Li Shen, and Gang Sun. Squeeze-and-excitation networks. In Proceedings of the IEEE conference on computer vision and pattern recognition, pp. 7132–7141, 2018.

\bibitem{AndrewGHowardMenglong}
Andrew G Howard, Menglong Zhu, Bo Chen, Dmitry Kalenichenko, Weijun Wang, Tobias Weyand, Marco Andreetto, and Hartwig Adam. Mobilenets: Efficient convolutional neural networks for mobile vision applications. arXiv preprint arXiv:1704.04861, 2017

\end{thebibliography}
\end{document}